\documentclass[journal]{IEEEtran}
\ifCLASSINFOpdf
  \usepackage[pdftex]{graphicx}
\else
\fi
\usepackage{amsmath}
\usepackage{makecell}
\usepackage{tabularx}
\ifCLASSOPTIONcompsoc
  \usepackage[caption=false,font=normalsize,labelfont=sf,textfont=sf]{subfig}
\else
  \usepackage[caption=false,font=footnotesize]{subfig}
\fi
\begin{document}
\title{Face Alignment Using K-cluster Regression Forests With Weighted Splitting } 

\author{Marek~Kowalski
        and Jacek~Naruniec

\thanks{Copyright (c) 2015 IEEE. Personal use of this material is permitted. However, permission to use this material for any other purposes must be obtained from the IEEE by sending a request to pubs-permissions@ieee.org.}        
\thanks{Marek Kowalski and Jacek Naruniec were with the Institute of Radioelectronics and Multimedia Technology, Warsaw University of Technology (e-mail: m.kowalski@ire.pw.edu.pl; j.naruniec@ire.pw.edu.pl).}% <-this % stops a space
}


\maketitle

\begin{abstract}
In this work we present a face alignment pipeline based on two novel methods: weighted splitting for K-cluster Regression Forests and 3D Affine Pose Regression for face shape initialization.
Our face alignment method is based on the Local Binary Feature framework, where instead of standard regression forests and pixel difference features used in the original method, we use our K-cluster Regression Forests with Weighted Splitting (KRFWS) and Pyramid HOG features. 
We also use KRFWS to perform Affine Pose Regression (APR) and 3D-Affine Pose Regression (3D-APR), which intend to improve the face shape initialization. APR applies a rigid 2D transform to the initial face shape that compensates for inaccuracy in the initial face location, size and in-plane rotation. 3D-APR estimates the parameters of a 3D transform that additionally compensates for out-of-plane rotation. 
The resulting pipeline, consisting of APR and 3D-APR followed by face alignment, shows an improvement of 20\% over standard LBF on the challenging IBUG dataset, and state-of-the-art accuracy on the entire 300-W dataset.
\end{abstract}


% \fi
\IEEEpeerreviewmaketitle

\section{Introduction}

\IEEEPARstart{F}{ace} alignment, also known as facial landmark localization, is an essential step in many computer vision methods such as face verification \cite{Bayes} and facial motion capture \cite{Animation}. The majority of face alignment methods proposed in the last several years are based on Cascaded Shape Regression (CSR), which was first proposed in \cite{ESR}.

CSR is usually initialized by placing the average face shape in the location provided by the face detector. Starting from this initialization, the face shape is refined in a fixed number of iterations. In each iteration features are extracted from the regions around each landmark of the face shape. The extracted features are then used in a regression method to estimate a correction to the current positions of the landmarks. 

In \cite{LBF} Ren et al. proposed Local Binary Features (LBF) where binary descriptors are created based on regression forests \cite{forests} built on simple pixel difference features. The proposed features were incorporated into the CSR framework, which lead to a very fast method, running at the speed of 3000 frames per second on standard desktop hardware. 

Our face alignment method is based on LBF, but instead of standard regression forests with pixel features, we use novel K-cluster Regression Forests with Weighted Splitting (KRFWS) and Pyramid HOG (PHOG) features. PHOG features consist of HOG \cite{HOG} taken over the same area with progressively larger cell sizes, more details can be found in \cite{KRF}.

K-cluster Regressions Forests (KRF) were originally proposed in \cite{KRF} for head pose estimation and car orientation estimation. In KRF, at each split, the target space is divided into K clusters, which locally minimize the loss function of the tree. Each K-fold split is performed using a separate k-class linear SVM classifier. The advantage of this type of splitting is that it allows to split on multidimensional features such as HOG, whereas standard regression forest splitting uses only scalar, one dimensional features.

In this work we propose K-cluster Regression Forests with Weighted Splitting (KRFWS) by introducing weights on individual samples for classification at each split. We test KRFWS on the Pointing'04 dataset \cite{Pointing}, for which KRF was originally designed, and show improved accuracy. 

Recently Li et al. proposed Affine-Transformation Parameter Regression for Face Alignment (APR) \cite{APR}, which estimates a 2D affine transform that refines the current face shape estimate. As opposed to the standard regression step in CSR, which refines the positions of the individual landmarks, APR applies a rigid transform. In \cite{APR} it was shown that embedding such alignment before and between standard cascaded shape regression iterations improves face alignment accuracy. 

In this work we propose two major extensions to the APR method. First, we show that using KRFWS instead of linear regression improves the accuracy of the estimated transform. Second, we propose 3D-APR, where, we estimate a 3D transform of the current face shape estimate. The use of a 3D transform allows for the compensation of out-of-plane rotation of the head, which is not possible with standard APR.

\begin{figure}[!t]
\centering
\subfloat[]{\includegraphics[width=0.32\linewidth]{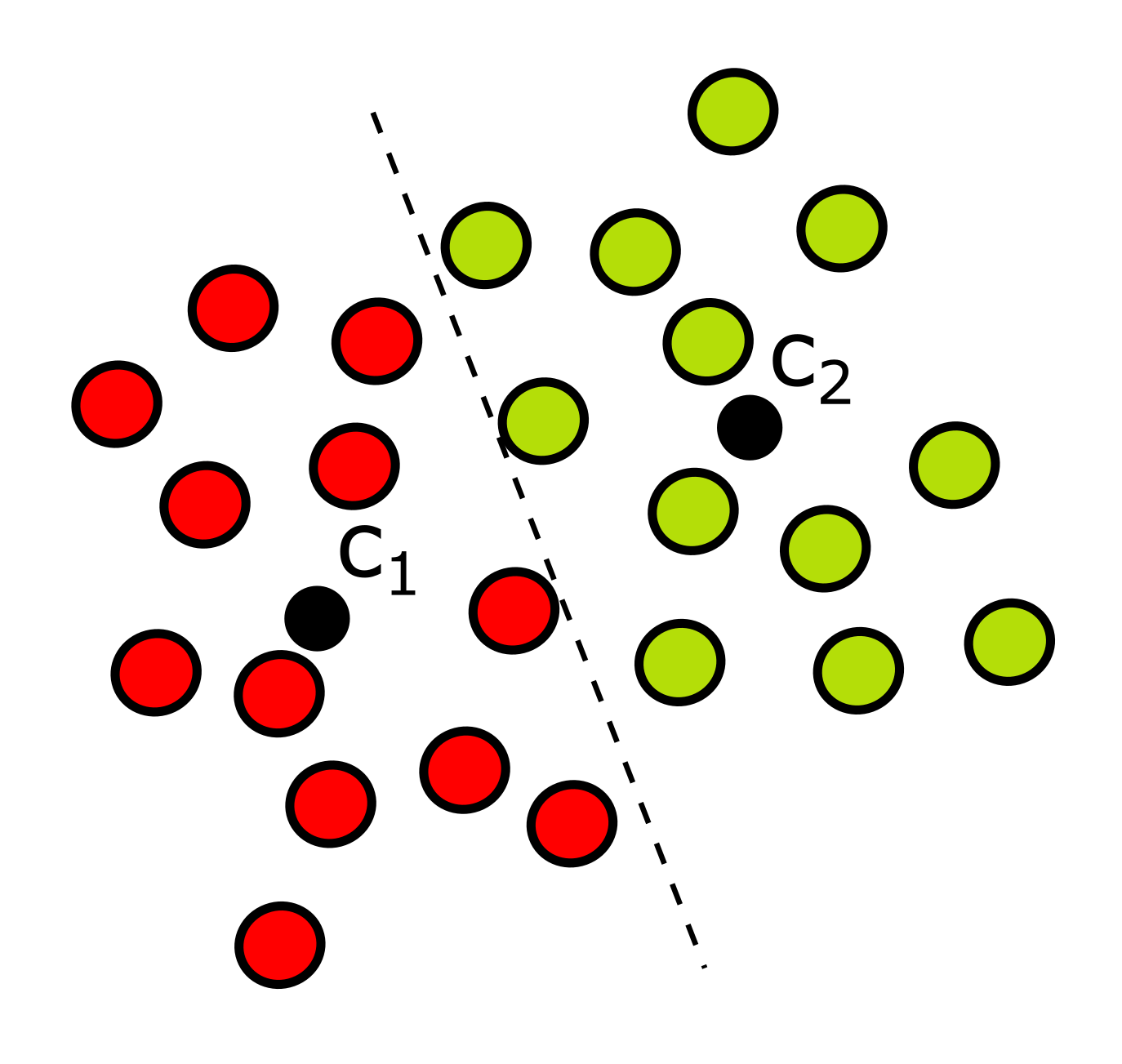}}
\quad
\subfloat[]{\includegraphics[width=0.32\linewidth]{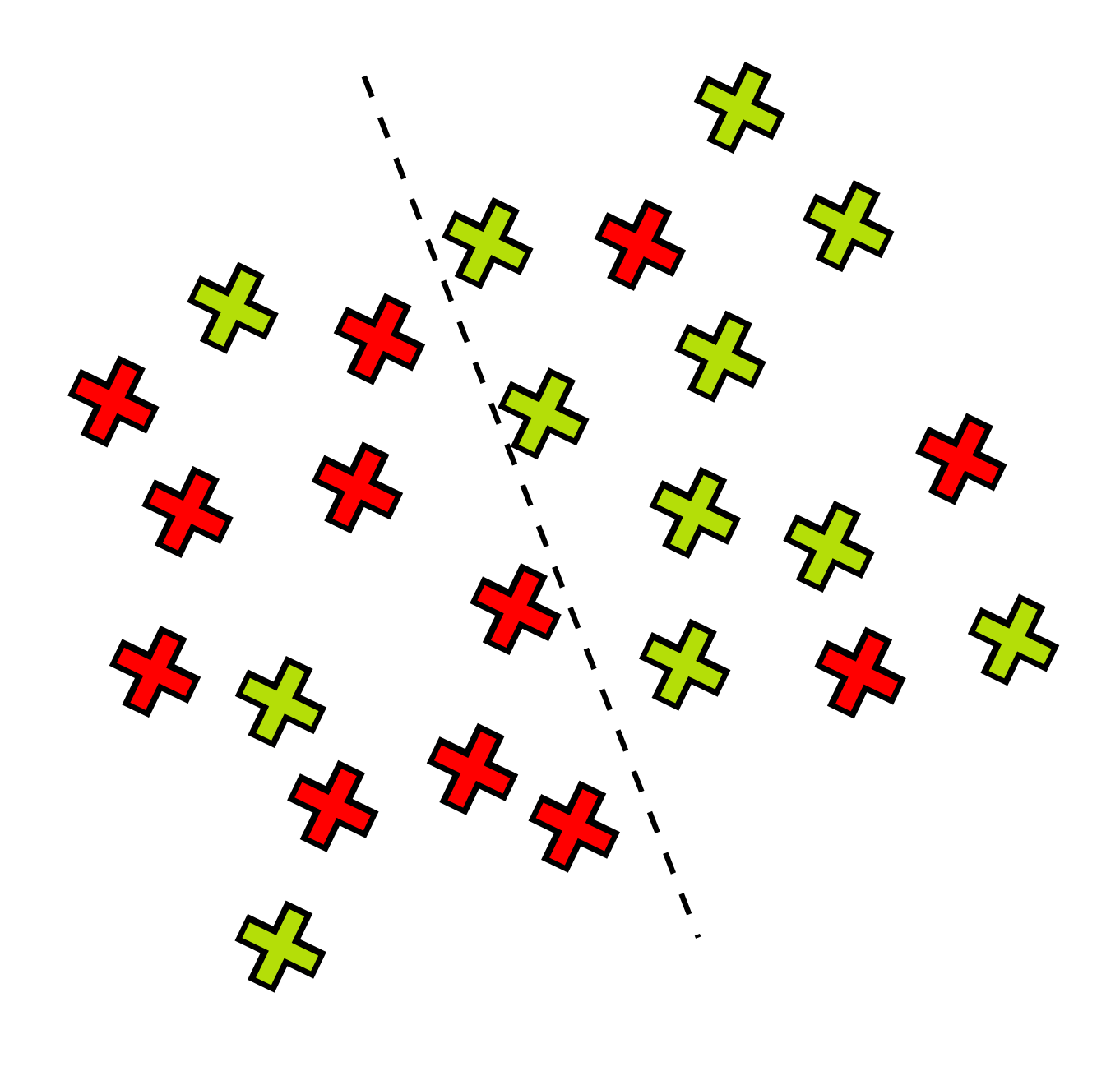}}\\
\subfloat[]{\includegraphics[width=0.32\linewidth]{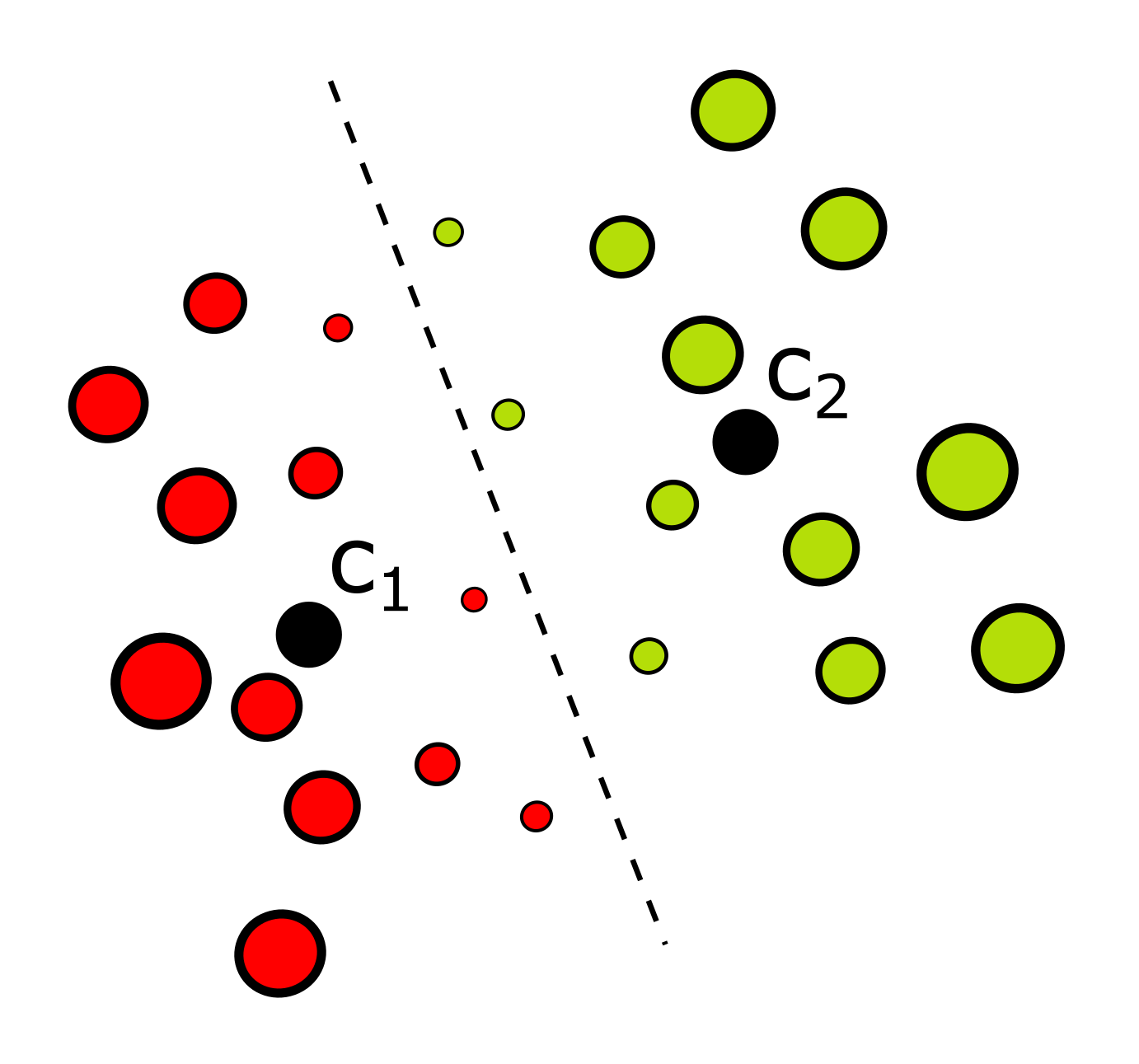}}
\quad
\subfloat[]{\includegraphics[width=0.33\linewidth]{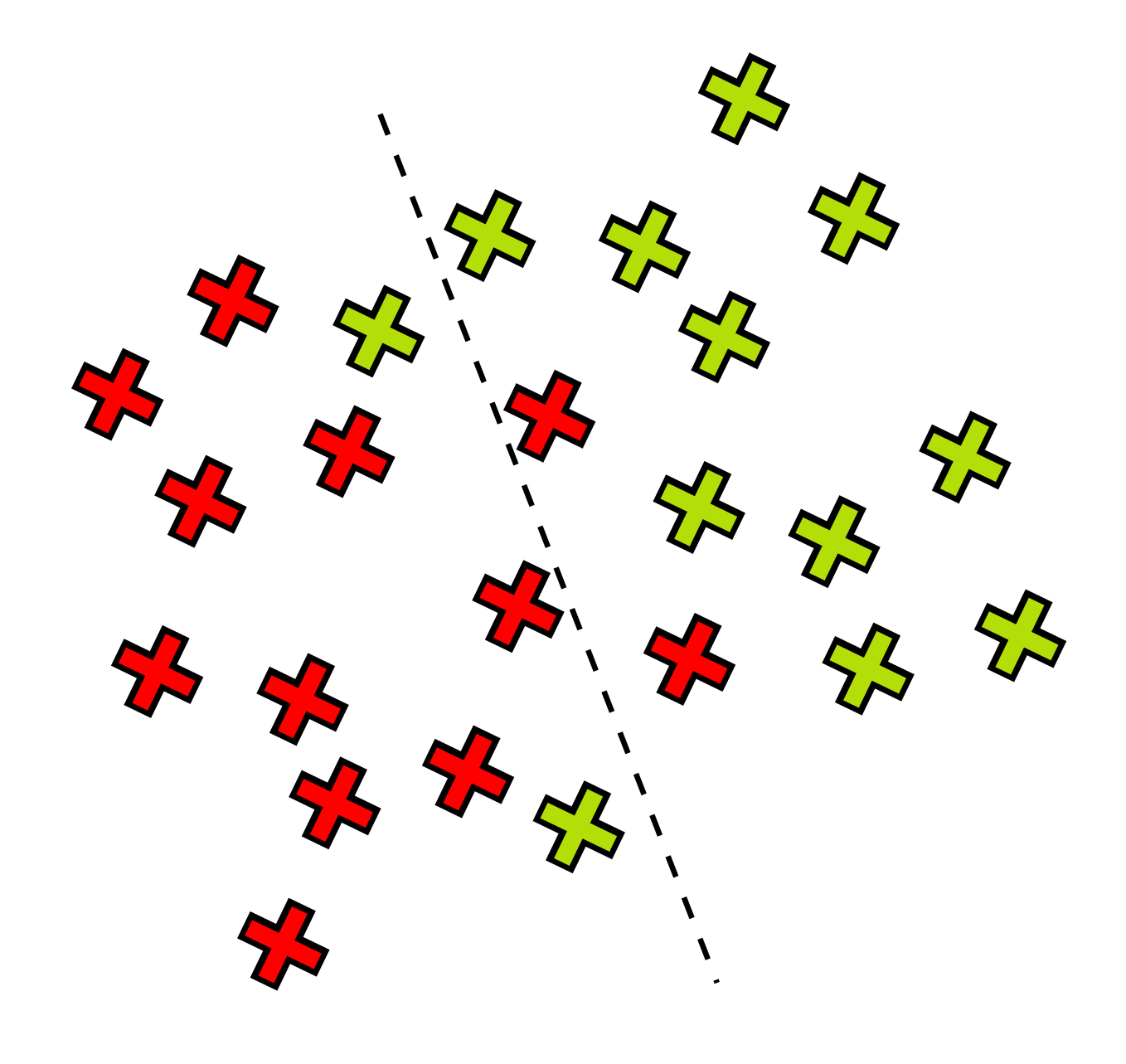}}
\caption{An illustration of the weighting scheme used in KRFWS and its influence on classification results. Figures (a) and (c) show a set of samples in a 2 dimensional target space $y$, divided into two clusters using k-means clustering. The size of each point corresponds to its weight, (a) uses uniform weighting, while (c) uses the weighting scheme proposed in KRFWS. Figures (b) and (d) show classification results obtained from data in (a) and (c). The misclassified samples in (d) are more likely to be located close to the border between the two clusters and in consequence their influence on SSE is limited.}
 \label{fig:drawing}
\end{figure}

\section{Methods}
\subsection{K-cluster Regression Forests with Weighted Splitting}
In this chapter we describe K-cluster Regression Forests with Weighted Splitting (KRFWS) which are an extension of K-cluster regression forests (KRF). Regression forests, like all other regression methods, map a point in the input space $\pmb{X}$ onto a point in the target space $\pmb{Y}$. KRF differ from other similar methods in the way splitting is performed and in the fact that in KRF each node has $K \geq 2$ child nodes, whereas in most methods each node has 2 children.

The aim of splitting is to find a rule that once applied to the input space $\pmb{X}$, partitions the target space $\pmb{Y}$ in a way that minimizes the loss function of the tree. KRF use the most common loss function, which is the sum of squared errors (SSE) of the child nodes. As explained in \cite{KRF}, the SSE of the child nodes of any given node is locally minimized by a partition defined by the result of k-means clustering on the target space of samples in that node.

In KRF, the problem of splitting is defined as a classification problem. To preserve the partition defined by the k-means clustering, a classifier is trained at each split. This is achieved by applying a $K$ class one-versus-rest linear SVM classifier on the input space of samples in each node. The advantage of this approach is that the splits can be performed on multidimensional features. This in turn facilitates the use of image features, for example PHOG, as the input space.

In most cases the separating hyperplane produced by the SVM does not partition the input set perfectly. In practice this means that some of the samples will not be forwarded to the correct child node. The misclassified samples increase the value of the loss function and make the splitting problem more difficult in the child nodes.

We propose to reduce the impact of this problem by giving more importance to the samples that, should they be misclassified, would have the largest impact on SSE. This is achieved by assigning each sample $i$ in the current node $T_j$ a weight $w_i\in [0, 1]$, where 0 would signify that the sample is of no importance. The weights are then used as an input to a weighted SVM classifier included in the LIBLINEAR \cite{LIBLINEAR} package. Below we describe our method for generating the sample weights for splits with $K=2$ clusters. While the method does generalize to any number of clusters, the description of the generalized procedure is beyond the scope of this short article. 

Given a node $T_j$ divided into two clusters $C_1$ and $C_2$ with their respective centroids $c_1$ and $c_2$, the weights are defined as follows:
\begin{equation} \label{eq:projection}
v_i = \left(y_i - \frac{c_2 + c_1}{2}\right)^T \cdot  \frac{c_2 - c_1}{\lVert c_2 - c_1 \rVert}  ,
\end{equation}
\begin{equation} \label{eq:normalization}
w_i = \frac{\lvert v_i \rvert }{\max_{i\in T_j}[\lvert v_i \rvert]},
\end{equation}
where $i \in T_j = C_1 \cup C_2$ and $y_i\in\pmb{Y}$. Equation \eqref{eq:projection} transforms each point $y_i$ of the target space into a distance $v_i$ of that point from the hyperplane separating the two clusters. The distances are then normalized in equation \eqref{eq:normalization} so that $w_i\in [0, 1]$.

Effectively each weight is the normalized distance of a given sample from the hyperplane separating $C_1$ and $C_2$. The samples close to the boundary between the two clusters have a small weight as even if they are incorrectly classified, the SSE will not increase greatly. The samples that are far from the boundary on the other hand, have large weights as, should they be misclassified, the SSE would be heavily affected. The weighting scheme and its influence are illustrated in Figure \ref{fig:drawing}.

KRFWS is not the only decision forest method that utilizes sample weights, one other method is boosting with decision trees \cite{boosting}. There are however several key differences between the two methods. In boosting with decision trees, the learning of each tree is dependent upon its predecessors. The samples are weighted based on their error in previously trained trees. In KRFWS each tree is trained independently using a different subset of the training data (bagging). The sample weights are calculated independently at each split, and reflect the influence a given samples would have on the loss if it was misclassified.

\subsection{Affine Pose Regression}
Affine Pose Regression (APR) was recently proposed in \cite{APR} as a method for improving the performance of face alignment methods. In contrast to Cascaded Shape Regression (CSR), APR estimates a rigid transforms of the entire face shape:
\begin{equation} \label{eq:transform}
S' = 
\begin{bmatrix}
a & b\\
c & d
\end{bmatrix}
S + 
\begin{bmatrix}
t_x & \dots & t_x\\
t_y & \dots & t_y
\end{bmatrix},
\end{equation}
where $S$, a $2\times n$ matrix, is the current estimate of the face shape, $n$ is the number of landmarks in the face shape and $a, b, c, d, t_x, t_y$ are the parameters of the transform. The parameters are estimated by linear regression based on HOG features extracted at the facial landmarks. APR can be applied before CSR or in between CSR iterations to efficiently compensate for inaccurate initialization of the face shape in scale, translation and in-plane rotation. 

In this work we propose to improve the original APR framework by using KRFWS instead of linear regression. We estimate all the transform parameters by creating separate KRFWS models for $a,b,c,d$ and a joint model for $t_x, t_y$. Instead of extracting features at individual landmarks we extract a single feature that covers the entire face. We show the effectiveness of our approach in experiments on the 300-W dataset \cite{300-W} in section \ref{sec:experiments}. 

\subsection{3D Affine Pose Regression}
As mentioned in the previous section, APR can be applied before CSR to compensate for inaccuracy in scale, translation and in-plane rotation of the face shape estimate. In this section we propose a method to extend APR by taking into account out-of-plane rotation of the head, namely: yaw and pitch. Our method, which we call 3D-APR, fits an average 3D face shape to the face in the image and uses the 2D projection of that shape as an initialization for face alignment. 

3D Affine Pose Regression (3D-APR) consists of two steps: first we fit an average 3D face shape $\bar{S}$ to the initial face shape estimate $S$. The fitting is accomplished using a scaled orthographic projection:
\begin{equation} \label{eq:scaledorto}
\bar{s} = k \cdot P \cdot \bar{S} + 
\begin{bmatrix}
t_x & \dots & t_x\\
t_y & \dots & t_y
\end{bmatrix},
\end{equation}
where $\bar{s}$ is the projected shape, $k$ is a scaling factor, $P$ are the first two rows of a rotation matrix and $t_x$, $t_y$ are translation parameters. The values of the parameters $\Gamma=\{k,P,t_x,t_y\}$ for any shape are obtained by solving the following optimization problem:
\begin{equation} \label{eq:gauss}
\newcommand{\argmin}{\operatornamewithlimits{arg\min}}
\Gamma = \argmin_{\Gamma} \lVert \bar{s} - S \rVert^2.
\end{equation}
The optimization is performed using the Gauss-Newton method as in \cite{2Dvs3D}. 

In the second step we estimate an update $\Delta$ to $\Gamma$ that refines the projected 3D shape $\bar{s}$ so that it is closer to the true shape of the face in the image. As in APR the estimation is performed using KRFWS based on a PHOG descriptor extracted at the face region. Separate KRFWS models are trained to estimate $k$ and $P$, a joint model is used for $t_x$ and $t_y$. The projection matrix $P$ is parametrized using euler rotations. In practice, the estimation of $P$ is equivalent to head pose estimation. 

Learning is performed similarly to learning in APR and CSR. The training set consists of a set of images with corresponding ground truth landmark locations. For each image a number of initial shapes are generated from the ground truth shape. For each initial shape, the initial parameters $\Gamma$ and the ground truth parameters $\Gamma'$ are obtained using equation \eqref{eq:gauss} and the ground truth annotations. KRFWS learning is then applied to map the PHOG descriptor to the update $\Delta = \Gamma' - \Gamma$.

\section{Experiments} \label{sec:experiments}

In this section we test the effectiveness of the proposed methods in affine pose regression, face alignment and head pose estimation. The parameters we use for our methods in APR and face alignment have been established through cross-validation, with the exception of the number of children $K$. $K$ was set following \cite{KRF}, where the authors have found $K=2$ to be optimal for a target space similar to ours. We plan to investigate different values of $K$ in future experiments..

\subsection{Affine pose regression}

We test the effectiveness of APR and 3D-APR on the 300-W dataset \cite{300-W}, which consists of face images with corresponding ground truth annotations of 68 characteristic points and bounding boxes generated by a face detector. The images in 300-W are gathered from several other datasets: AFW \cite{AFW}, HELEN \cite{HELEN}, IBUG \cite{300-W} and LFPW \cite{LFPW}. For learning we use the AFW dataset and the training subsets of the HELEN and LFPW datasets, which together consist of 3148 face images. Our test dataset consists of two subsets: the challenging IBUG dataset (135 images) and the less challenging test subsets of the LFPW and HELEN datasets (554 images). Together the two datasets form what we refer to as the full set. This division of the 300-W dataset is a standard in face alignment testing, employed in many recent articles \cite{LBF, CFSS, TransferredDCNN}.

Each method is initialized with the face detector bounding box provided in the 300-W dataset. Similarly to \cite{LBF},\cite{CFSS} we use the inter-pupil distance normalized landmark error, all errors are expressed as the \% of the inter-pupil distance. The pupil locations are assumed to be the centroids of the landmarks located around each of the eyes.

Five different configurations of APR and 3D-APR are tested: (1) Linear APR with feature extraction at landmarks, (2) Linear APR with a single feature extracted at the face center, (3) KRF APR with a single feature extracted at the face center, (4) KRFWS APR with a single feature extracted at the face center, (5) KRFWS APR followed by 3D-APR with a single feature extracted at the face center (Combined APR, CAPR). In all of the configurations the images are rescaled so that the face size is approximately $64\times 64$ pixels. In all experiments APR is performed for two iterations, while 3D-APR is performed once.

In the first configuration Pyramid HOG \cite{KRF} features covering $32\times 32$ pixels are extracted at each landmark. The input descriptor for APR is formed by concatenating the descriptors from each of the landmarks.

In configurations (2), (3), (4) and (5) a single PHOG is extracted at the face center. As the feature size is not a concern in this scenario (only one feature is extracted instead of 68) we use the extended version of the HOG feature described in \cite{extHOG}. The descriptor covers an area of $64\times 64$ pixels.

The results of the experiments are shown in Table \ref{tab:APR}. KRFWS APR outperforms Linear APR on the challenging subset by 6\%. CAPR shows the best accuracy of all tested methods, reducing the error of Linear APR by 35\% on the full set.

\begin{table}[htbp]
\caption{Error of APR methods on the 300-W dataset.} \label{tab:APR}
\begin{tabularx}{\linewidth}{ >{\centering\arraybackslash}X c c c }
\Xhline{4\arrayrulewidth}
Methods & \makecell{Common \\ subset} & \makecell{Challenging \\ subset} & Full set\\
\hline	
Linear APR & 12.70 & 26.00 & 15.29 \\
Linear APR single feature & 12.77 & 25.85 & 15.32 \\
KRF APR & 11.48 & 24.80 & 14.08 \\
KRFWS APR & 11.37 & 24.28 & 13.88 \\
KRFWS APR + 3D-APR (CAPR) & 8.61 & 15.26 & 9.90 \\
\Xhline{4\arrayrulewidth}
\end{tabularx}
\end{table}

\subsection{Face alignment}
\begin{figure}[!t]
\centering
\includegraphics[width=0.93\linewidth]{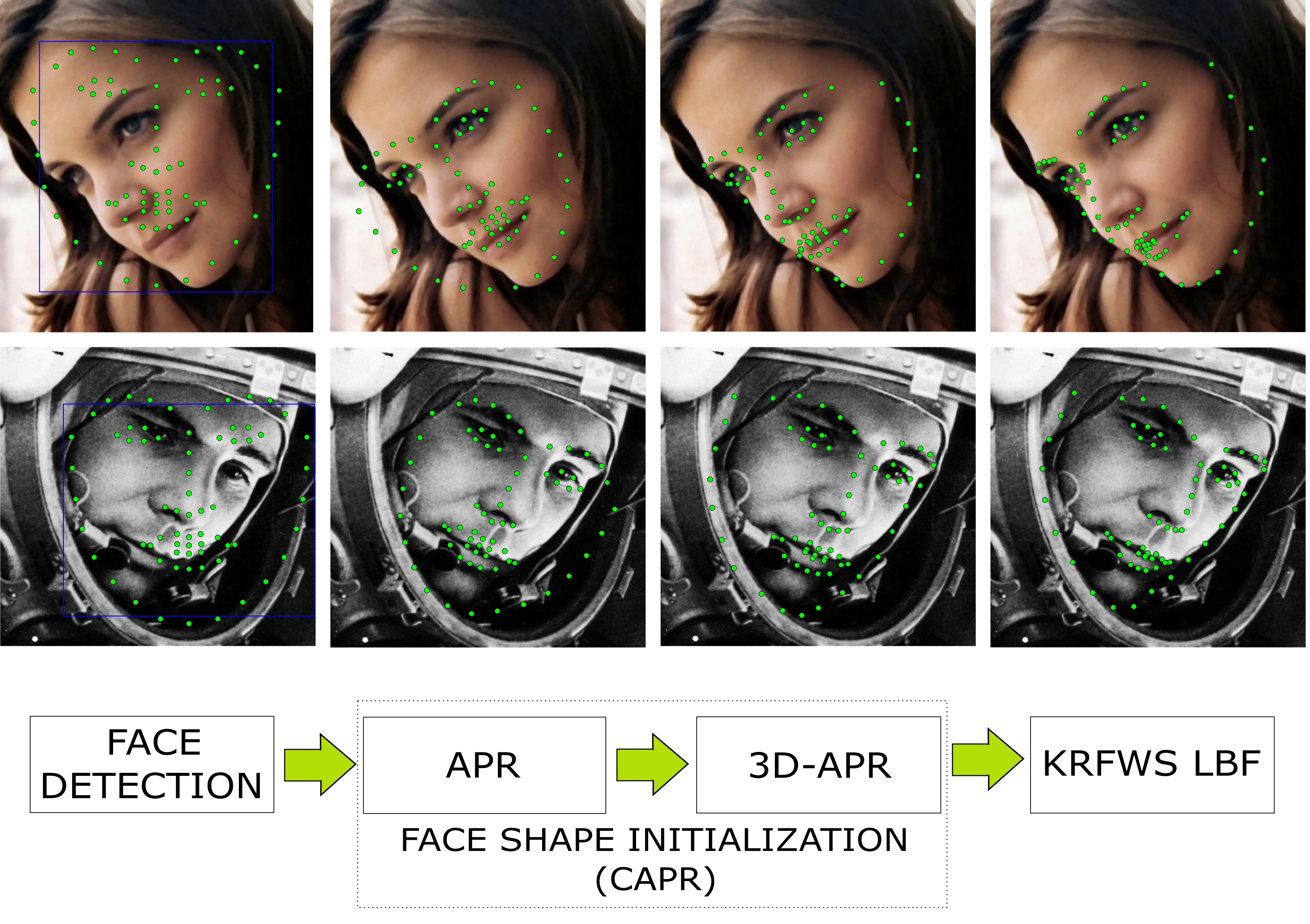}\\
\caption{A diagram showing the proposed face alignment pipeline. The images above, taken from the IBUG dataset, show the results at the consecutive stages of the pipeline. }
 \label{fig:faces}
\end{figure}

In face alignment we use the same training and evaluation data as in the APR experiments. In order to facilitate comparison with other methods we report the results of our full pipeline for both inter-pupil normalisation \cite{LBF},\cite{CFSS} and inter-ocular normalisation \cite{MDM}. Our face alignment method uses the Local Binary Feature framework \cite{LBF}, where instead of standard regression forests we use KRFWS, and instead of pixel difference features we use PHOG. The forest generated for each landmark consists of 5 trees with a maximum depth of 7, the PHOG extracted at landmarks cover an area of $32\times 32$ pixels each, with a single block per pyramid level. Similarly to \cite{LBF}, our method is performed for 5 iterations.

We test the accuracy of our face alignment method with four different initialization configurations: (1) face detector bounding boxes provided in the 300-W dataset, (2) Linear APR, (3) KRFWS APR, (4) KRFWS APR + 3D-APR (CAPR). 

The results of the experiments are shown in Table \ref{tab:alignment} along with the results of several state-of-the-art methods, all trained on the same datasets and all initialized as in configuration (1). The methods we compare to perform face alignment using a variety of machine learning tools, including: linear regression \cite{SDM}, decision trees \cite{LBF} and deep learning \cite{TransferredDCNN, MDM}.

The experiments we have performed show the effectiveness of our face alignment combined with KRFWS APR and 3D-APR. In configuration (1) the proposed face alignment method outperforms the original LBF in all cases by between 2\% and 8\%. The addition of KRFWS APR and CAPR for initialization leads to an improvement of 18\% and 20\% over the original LBF on the challenging IBUG dataset.

Our face alignment combined with CAPR initialization shows state-of-the-art results on all datasets. The addition of CAPR resulted in an error reduction of over 13\% on the challenging subset in comparison to face detector initialization.

Figure \ref{fig:faces} shows a diagram of our face alignment pipeline along with images at its consecutive stages.

\begin{table}[htbp]
\caption{Error of face alignment methods on the 300-W dataset.} \label{tab:alignment}
\begin{tabularx}{\linewidth}{ >{\centering\arraybackslash}X c c c }
\Xhline{4\arrayrulewidth}
Methods & \makecell{Common \\ subset} & \makecell{Challenging \\ subset} & Full set\\
\hline
\multicolumn{4}{c}{inter-pupil normalisation} \\
\hline
ESR \cite{ESR} & 5.28 & 17.00 & 7.58 \\

SDM \cite{SDM} & 5.60 & 15.40 & 7.52 \\

LBF \cite{LBF} & 4.95 & 11.98 & 6.32 \\

Transferred DCNN \cite{TransferredDCNN} & 4.73 & 12.37 & 6.23\\

CFSS \cite{CFSS} & 4.73 & 9.98 & 5.76 \\

\hline	
\textbf{KRFWS LBF} & \textbf{4.84} & \textbf{10.96} & \textbf{6.03} \\
\textbf{Linear APR + KRFWS LBF} & \textbf{4.76} & \textbf{10.57} & \textbf{5.89} \\
\textbf{KRFWS APR + KRFWS LBF} & \textbf{4.65} & \textbf{9.82} & \textbf{5.66} \\
\textbf{CAPR + KRFWS LBF} & \textbf{4.62} & \textbf{9.48} & \textbf{5.57} \\
\hline
\multicolumn{4}{c}{inter-ocular normalisation} \\
\hline
MDM \cite{MDM} & - & - & 4.05\\
\textbf{CAPR + KRFWS LBF} & \textbf{3.34} & \textbf{6.56} & \textbf{3.97} \\
\Xhline{4\arrayrulewidth}
\end{tabularx}

\end{table}

\subsection{Head pose estimation} \label{sec:headpose}
In order to compare KRFWS to the original KRF we test our method on the head pose estimation task. We use the Pointing'04 \cite{Pointing} dataset, which was also used to test the original KRF. The Pointing'04 dataset consists of images of 15 subjects, each photographed in two separate sessions. During both sessions each subject had 93 photographs taken with pitch and yaw of the head both varying from -90$^{\circ}$ to +90$^{\circ}$. Each image in the dataset is accompanied by a manually annotated bounding box containing the head. \looseness=-1

For fair comparison we use experimental settings identical to those proposed in \cite{KRF}. From each image we extract a single Pyramid HOG feature, set K in KRFWS to 2 and have each forest consist of 20 trees. The splitting stops when a node has less than 5 samples.

We compare our method to the baseline KRF in two experiments, in both we use the Mean Absolute Error (MAE) measure. In the first experiment we perform 2-fold cross validation, where each fold consists only of images from a single session. In the second experiment we perform 5-fold cross validation, where the images in each fold are chosen at random. Our method shows an improvement of 4\% over the baseline KRF in the first test and a 3.7\% improvement in the second test. The results of both tests can be found in Table \ref{tab:pose} along with the results for KRF and Adaptive KRF (AKRF) \cite{KRF}. \looseness=-1

In MATLAB the training time for a single tree in the 5-fold cross validation test is 6.65 sec for KRF and 6.45 sec for KRFWS. This shows that the training of KRFWS is actually faster than training of KRF, despite the additional weight calculation step. We believe that this is because in KRFWS less splits are required to reach the stopping criterion.

\begin{table}[htbp] 
\centering
\caption{Mean Absolute Error values for 2-fold and 5-fold cross validation on the Pointing'04 dataset.} \label{tab:pose}
\begin{tabular}{ | c || c | c | c | }
\hline	
Method & yaw & pitch & average \\
\hline	
\multicolumn{4}{|c|}{2-fold cross validation} \\
\hline	
\textbf{KRFWS} & \textbf{4.95} & \textbf{3.35} & \textbf{4.15 $\pm$ 0.041} \\
\hline	
KRF (baseline) & 5.06 & 3.59 & 4.32 $\pm$ 0.138\\
\hline	
AKRF & 5.45 & 3.92 & 4.68 $\pm$ 0.072 \\
\hline	
\multicolumn{4}{|c|}{5-fold cross validation} \\
\hline	
\textbf{KRFWS} & \textbf{5.07} & \textbf{2.65} & \textbf{3.86 $\pm$ 0.177} \\
\hline	
KRF (baseline) & 5.13 & 2.88 & 4.01 $\pm$ 0.171\\
\hline	
AKRF & 5.57 & 3.39 & 4.48 $\pm$ 0.267 \\
\hline	
\end{tabular}
\end{table}

\section{Conclusion}
In this article we have proposed a face alignment pipeline based on novel K-cluster Regression Forests with Weighted Splitting. Our pipeline consists of two separate stages: face shape initialization and face alignment. The first step performs APR and novel 3D-APR to improve the initial shape, provided by the face detector, in terms of translation, scale and in and out of plane rotation. The second step performs face alignment using an adapted version of the LBF framework. The proposed face alignment pipeline shows state-of-the-art results on the entire 300-W dataset.

\ifCLASSOPTIONcaptionsoff
  \newpage
\fi



\bibliographystyle{IEEEtran}
\bibliography{IEEEabrv,biblio}

\end{document}